\documentclass[letterpaper]{article} 
\usepackage{aaai2026}  
\usepackage{times}  
\usepackage{helvet}  
\usepackage{courier}  
\usepackage[hyphens]{url}  
\usepackage{graphicx} 
\usepackage{natbib}  
\usepackage{caption} 
\usepackage{algorithm}
\usepackage{algorithmic}
\usepackage{amssymb}
\usepackage{amsmath}
\usepackage{booktabs}
\usepackage{multirow}
\usepackage{adjustbox}
\usepackage{subcaption}
\usepackage{amsmath}       
\usepackage{amssymb}       
\usepackage{newfloat}
\usepackage{listings}
\DeclareCaptionStyle{ruled}{labelfont=normalfont,labelsep=colon,strut=off} 
\floatstyle{ruled}
\newfloat{listing}{tb}{lst}{}
\floatname{listing}{Listing}
\nocopyright 

\title{DReSS: Data-driven Regularized Structured
Streamlining for Large Language Models}
\author{
  Mingkuan Feng$^{1*}$,\quad  Jinyang Wu$^1$\thanks{Equal Contribution.},\quad  Shuai Zhang$^1$\thanks{Corresponding authors.},\quad  Ruihan Jin$^1$,\quad  \\ \textbf{Pengpeng Shao$^1$,\quad Feihu Che$^1$,\quad Zhengqi Wen$^1$, \quad Jianhua Tao$^{1\dagger}$}\\
}
\affiliations{
    \textsuperscript{\rm 1}Tsinghua University\\

}

\usepackage{bibentry}

\begin{document}

\maketitle

\begin{abstract}
Large language models (LLMs) have achieved significant progress across various domains, but their increasing scale leads to high computational and memory costs. Recent studies show that LLMs exhibit sparsity, which can be exploited for pruning. Existing pruning methods typically follow a prune-then-finetune paradigm. Since the pruned components still contain valuable information, their direct removal often leads to irreversible performance degradation, causing expensive fine-tuning to recover performance. To address this, we propose a new paradigm: first apply regularization, then prune, and finally fine-tune. Based on this paradigm, we propose DReSS, a simple and effective \textbf{D}ata-driven \textbf{Re}gularized \textbf{S}tructured \textbf{S}treamlining method for LLMs. By using a small amount of data to regularize the components before pruning, DReSS transfers the important information to the remaining parts of the model in advance. Compared to direct pruning, this can reduce the information loss caused by parameter removal, thereby enhancing its language modeling capabilities. We evaluate our method on various LLMs, including Phi-2, OPT, LLaMA2, LLaMA3. Experimental results demonstrate DReSS even without recovery fine-tuning (RFT) achieves comparable performance to previous methods, drastically alleviating computational costs. Moreover, DReSS significantly outperforms existing powerful pruning methods even under extreme pruning ratios, significantly reducing latency and increasing throughput.
\end{abstract}


\section{Introduction}
Large language models (LLMs) have achieved significant advancements across a wide range of tasks, demonstrating their robust capabilities~\cite{zhang2022opt,achiam2023gpt,touvron2023LLaMA,wu2024exampleshighlevelautomatedreasoning}. However, as the model size increases, the growing number of parameters leads to significant computational and memory requirements, which significantly hinder the practical deployment of LLMs. Consequently, it is critical to develop methods that can reduce model size while maintaining performance.

To address these challenges, several methods have been proposed, including pruning~\cite{frantar2023sparsegpt,sun2023simple,an2024fluctuation}, quantization~\cite{frantar2022gptq,xiao2023smoothquant}, knowledge distillation~\cite{shridhar-etal-2023-distilling,hsieh2023distilling}, and low-rank decomposition~\cite{saha2023matrix}. In this work, we mainly focus on pruning which is an efficient and highly generalizable approach that can be seamlessly integrated with other model compression strategies. Pruning techniques are generally classified into two primary categories: unstructured pruning~\cite{frantar2023sparsegpt,sun2023simple} and structured pruning~\cite{ma2023llm,an2024fluctuation}. Compared to unstructured pruning, structured pruning offers the flexibility to do recovery fine-tuning (RFT) for specific downstream tasks without relying on specialized hardware~\cite{zhu2024survey}. The model obtained through structured pruning typically achieves much faster inference speed due to the regular data patterns.

Despite these advancements, existing structured pruning methods still have limitations. They all follow the paradigm of first selecting channels or layers to prune based on a designed metric, and then performing RFT~\cite{chavan2024faster}. However, they neglect that important information can also exist in the pruned parts~\cite{dettmers2022gpt3,xiao2023smoothquant,yin2023outlier}, and directly removing them leads to an irreversible decline in model performance. Thus, the pruned models often require extensive data for RFT to recover performance~\cite{ma2023llm}. Additionally, high pruning ratios often cause performance collapse, limiting their effectiveness in acceleration.

\begin{figure*}[ht]
	\centering
\includegraphics[width=1.00\textwidth]{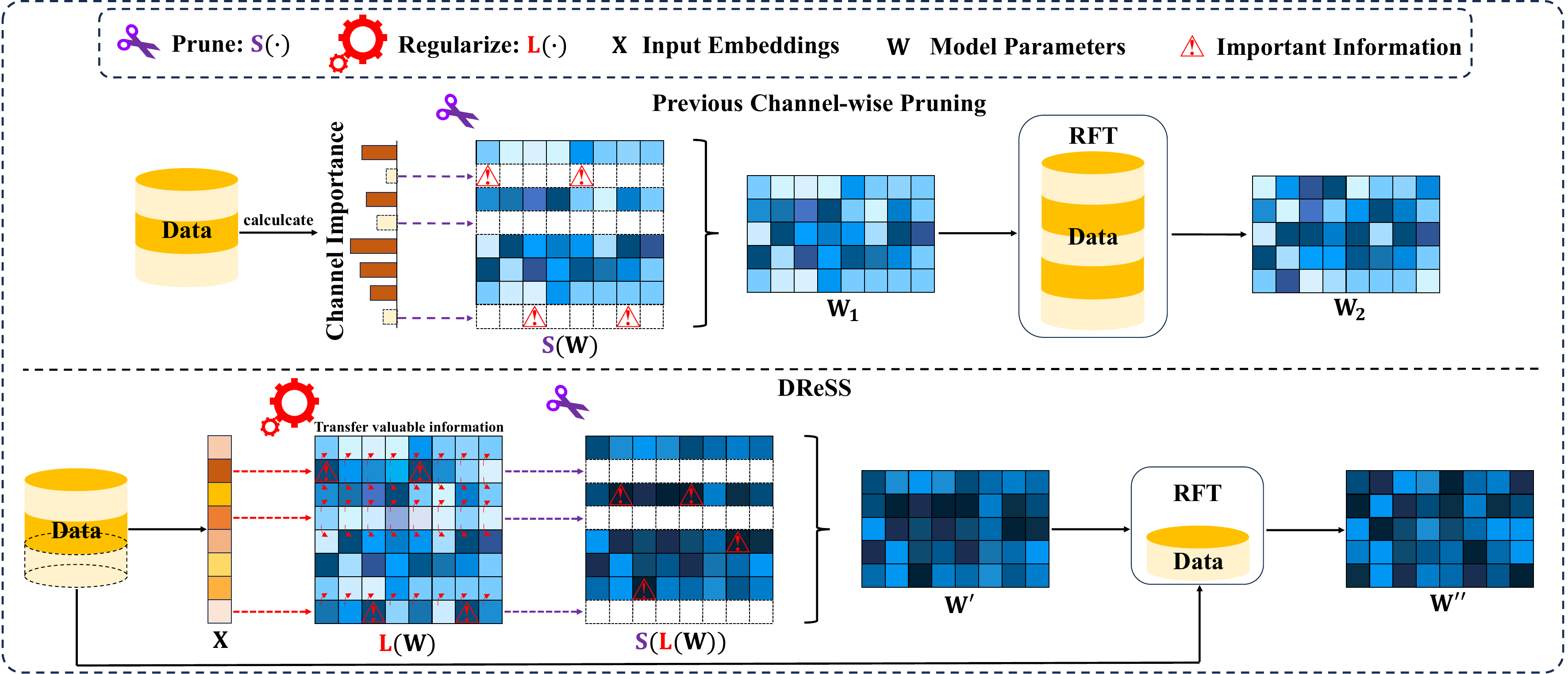}
    \caption{A comparison between previous channel-wise pruning methods and DReSS. Deeper blue square represents greater performance impact, the taller cylinder represents larger data volume. \textbf{Above}: Previous methods first select channels based on importance, followed by pruning and then do RFT. \textbf{Below}: DReSS first regularizes channels to transfer important information, prunes the channels, then do RFT with a small amount of data.}
  \label{picture1}
\end{figure*}

To tackle these limitations, we propose a new pruning paradigm that first applies regularization to the components to be pruned, then prunes those parameters, finally performs RFT. To the best of our knowledge, we are the first to propose this paradigm. The regularization process explicitly transfers important information from the parts to be pruned to the remaining parts, thereby reducing the information loss caused by direct parameter removal. Based on this paradigm, we introduce DReSS, a \textbf{D}ata-driven \textbf{Re}gularized \textbf{S}tructured \textbf{S}treamlining method for LLMs. The comparison of the DReSS with existing channel-wise pruning approaches is shown in Figure~\ref{picture1}. Specifically, the process of DReSS is divided into the following steps: (1) select data for pre-pruning regularization and post-pruning RFT, (2) apply regularization to the selected channels in the parameter matrices, (3) prune the regularized channels, finally (4) do RFT. \underline{\textit{First}}, we randomly select a small subset of data from widely used benchmark datasets. As the data size is small, the overhead of regularization and RFT remains minimal. \underline{\textit{Second}}, we determine the channels to be pruned based on the pruning ratio, and apply regularization ($\ell_1$-norm or $\ell_2$-norm) to these channels using the selected data. This redistributes important information to the unpruned parts, thereby significantly mitigating performance degradation caused by direct pruning. As a result, the model retains strong performance at high pruning ratios, enabling significant speedup. \underline{\textit{Third}}, we prune the regularized channels. \underline{\textit{Finally}}, we perform RFT on the pruned model using a small subset of samples selected in the first step. Extensive experiments show that DReSS significantly outperforms existing powerful pruning methods in perplexity and accuracy across various pruning ratios on different models and is able to provide considerable acceleration. The main contributions are summarized as follows:
\begin{itemize}
\item\textbf{Propose A New Paradigm: }By sequentially applying regularization ($\ell_1$-norm or $\ell_2$-norm), pruning, and RFT, DReSS minimizes information loss caused by direct parameter removal, thereby improving the model's language modeling capabilities.
\item\textbf{High Performance: }DReSS surpasses competitive structured pruning methods in perplexity and accuracy, achieving notable improvements in throughput and reduced latency compared to the dense models.
\item\textbf{Low Overhead: }By using only a small amount of data for regularization and optional RFT, DReSS incurs minimal overhead.
\end{itemize}
\section{Related Works}
\subsection{Pruning Methods}
Pruning redundant weights has been an effective way to reduce deep neural network complexity for decades.~\cite{lecun1989optimal,hassibi1993optimal,han2015learning}. Pruning methods can be broadly categorized into two types: unstructured pruning~\cite{kurtic2022optimal,zhang2024plug,xu2024besa} and structured pruning~\cite{xia2023sheared,gao2024disp}. 

Unstructured pruning methods remove individual weights based on designed metrics. Magnitude~\cite{han2015learning} removes smaller weights, Wanda~\cite{sun2023simple} considers both weights and activations, and SparseGPT~\cite{frantar2023sparsegpt} uses Hessian matrix. They require specialized hardware to accelerate~\cite{xia2023flash} and a high sparsity to achieve substantial acceleration~\cite{wang2020sparsert}.

Structured pruning methods, including channel-wise pruning~\cite{an2024fluctuation} and layer-wise pruning~\cite{men2024shortgpt}. Channel-wise pruning methods remove less important channels of the parameter matrices. For instance, SliceGPT~\cite{ashkboos2024slicegpt} prune channels with smaller eigenvalues. LLM Surgeon~\cite{van2023llm} periodically updates model weights and structures, pruning more aggressively in the initial layers. Layer-wise pruning methods, such as SLEB~\cite{song2024sleb}, iteratively prune transformer layers based on their importance. These methods have a key limitation: even less important channels or layers may contain valuable information, and pruning them directly often leads to information loss. To address this, we investigate the way of `shifting' important information from the pruned parts to the remaining parts, which could reduce information loss caused by pruning.
\subsection{Regularization}
Regularization is widely used in machine learning, such as feature selection~\cite{tibshirani1996regression} and preventing model overfitting~\cite{santos2022avoiding}. The $\ell_1$-norm drives certain coefficients to zero, while $\ell_2$-norm encourages smoother solutions~\cite{boyd2004convex}. Both of them can significantly alter the distribution pattern of the data~\cite{han2015learning,tao2023structured}. Motivated by this, we propose a new pruning paradigm, in which the regularization process can transfer important information from the pruned parameter space to the remaining parts of the model, thus reducing the information loss caused by parameter removal.
\section{Methodology}
\label{headings}
In this section, we provide a detailed introduction to DReSS. The overall DReSS algorithm is summarized in Figure~\ref{picture1} and Algorithm \ref{alg:example}, our method consists of four main steps: 
\begin{itemize}
\item\textbf{Data Selection: }Select a small amount of data for pre-pruning regularization and post-pruning RFT.
\item\textbf{Regularization: }Apply regularization to the selected channels of weight matrices to transfer important information from the pruned parts to the remaining parts, thereby reducing information loss caused by pruning and mitigating the resulting performance degradation.
\item\textbf{Pruing: }Remove the channels that were regularized in the previous step.
\item\textbf{RFT: }Perform RFT on the pruned model to compensate for the performance loss caused by pruning.
\end{itemize}
\subsection{Data Selection}
\label{3.1}
For fairness, the data is selected from widely used datasets. For instance, approximately 1,000 samples are randomly selected from the WikiText-2~\cite{merity2016pointer} training set for pre-pruning regularization and post-pruning RFT. 
\subsection{Regularization}
\label{3.2}
For clarity, we first define the notations. \( p \) is the pruning ratio, indicating \( p\% \) of model parameters will be pruned. The model dimension is \( d \), and the number of layers is \( l \). Additionally, \( \mathbf{W_{emb}} \in \mathbb{R}^{b \times n \times d} \) is the data feature map, and \( \mathbf{W_{pos}} \in \mathbb{R}^{b \times n \times d} \) is the positional feature map, where \( b \) is the batch size and \( n \) is the number of tokens. The model weights are represented by \( \mathbf{W} \in \mathbb{R}^{d_1\times d_2} \). Specifically, in the \( i \)th layer attention block, \( \mathbf{W_Q^i} \in \mathbb{R}^{d \times d_1} \) is query matrix, \( \mathbf{W_K^i} \in \mathbb{R}^{d \times d_1} \) is key matrix, \( \mathbf{W_V^i} \in \mathbb{R}^{d \times d_1} \) is value matrix, and \( \mathbf{W_o^i} \in \mathbb{R}^{d_1 \times d} \) is output matrix. In the \( i \)th layer FFN block, \( \mathbf{W_{up}^i} \in \mathbb{R}^{d \times d_2} \) is up projection matrix, and \( \mathbf{W_{down}^i} \in \mathbb{R}^{d_2 \times d} \) is down projection matrix. In the \( i \)th LayerNorm, \( \mathbf{\alpha^i} \in \mathbb{R}^d \) is scaling vector and \( \mathbf{\beta^i} \in \mathbb{R}^d \) is offset vector. \( \mathbf{W_{lm}} \in \mathbb{R}^{d \times b\times n} \) denotes the language modeling matrix. \( \mathbf{I} \in \mathbb{R}^{d \times d}\) is the identity matrix, let \( \mathbf{R} \in \mathbb{R}^{d \times d} \) is a pseudo-index selection matrix, which is a diagonal matrix consisting of \( (1 - p\%) \times d \) zeros and \( p\% \times d \) ones, where the ones correspond to the selected indices. \( \mathbf{R}\) is identical across all layers.

\begin{algorithm}[tb]
   \caption{DReSS algorithm.}
   \label{alg:example}
\begin{algorithmic}
   \STATE {\bfseries Input:} selected data $\mathbf{X}$, number of layers in LLMs $l$, initial model $\mathbf{W}$, norm type: $\text{flag}$.
\IF{flag==0} 
    \STATE \textbf{continue}\hfill \text{// $\ell_2$-norm}
\ELSIF{flag==1} 
    \STATE use Proposition 1 to transform problem \hfill \text{// $\ell_1$-norm}
\ENDIF
   \STATE$\mathbf{X_1}, \mathbf{X_2}\leftarrow divide(\mathbf{X})$
   \STATE $\tilde{\mathcal{L}}_{\text{1}} \leftarrow regularize(\mathbf{W_{emb}}, \mathbf{W_{pos}}, \mathbf{X_1})$
   \STATE$\tilde{\mathcal{L}}_{\text{att}} \leftarrow 0$, $\tilde{\mathcal{L}}_{\text{FFN}} \leftarrow 0$, $\tilde{\mathcal{L}}_{\text{2}} \leftarrow 0$
   \FOR{$i=1$ {\bfseries to} $l$}
     \STATE $\tilde{\mathcal{L}}_{\text{att}} \leftarrow \tilde{\mathcal{L}}_{\text{att}} + regularize(Attention, \mathbf{X_1})$
     \STATE $\tilde{\mathcal{L}}_{\text{FFN}} \leftarrow \tilde{\mathcal{L}}_{\text{FFN}} + regularize(FFN,\mathbf{X_1})$
     \STATE $\tilde{\mathcal{L}}_{\text{2}} \leftarrow \tilde{\mathcal{L}}_{\text{2}} + regularize(LayerNorm, \mathbf{X_1})$
   \ENDFOR
   \STATE $\tilde{\mathcal{L}}_{\text{3}} \leftarrow regularize(\mathbf{W_{lm}}, \mathbf{X_1})$
   \STATE $\tilde{\mathcal{L}}_{\text{remain}} \leftarrow \tilde{\mathcal{L}}_{\text{1}} + \tilde{\mathcal{L}}_{\text{2}} + \tilde{\mathcal{L}}_{\text{3}}$
   \STATE $\mathcal{L}(\mathbf{W}, \mathbf{X}) \leftarrow forward(\mathbf{W}, \mathbf{X_1})$
   \STATE $\mathcal{L}_{\text{sum}} \leftarrow \mathcal{L}(\mathbf{W}, \mathbf{X}) + \tilde{\mathcal{L}}_{\text{att}} + \tilde{\mathcal{L}}_{\text{FFN}} + \tilde{\mathcal{L}}_{\text{remain}}$
   \STATE update $\mathbf{W}$ using backpropagation algorithm
   \STATE $\mathbf{W} \leftarrow Prune(\mathbf{W}, \mathbf{R})$
   \STATE $\mathbf{W} \leftarrow Optional\_RFT(\mathbf{W}, \mathbf{X_2})$
\end{algorithmic}
\end{algorithm}
\begin{figure}[ht]
\begin{center}
\centerline{\includegraphics[width=\columnwidth]{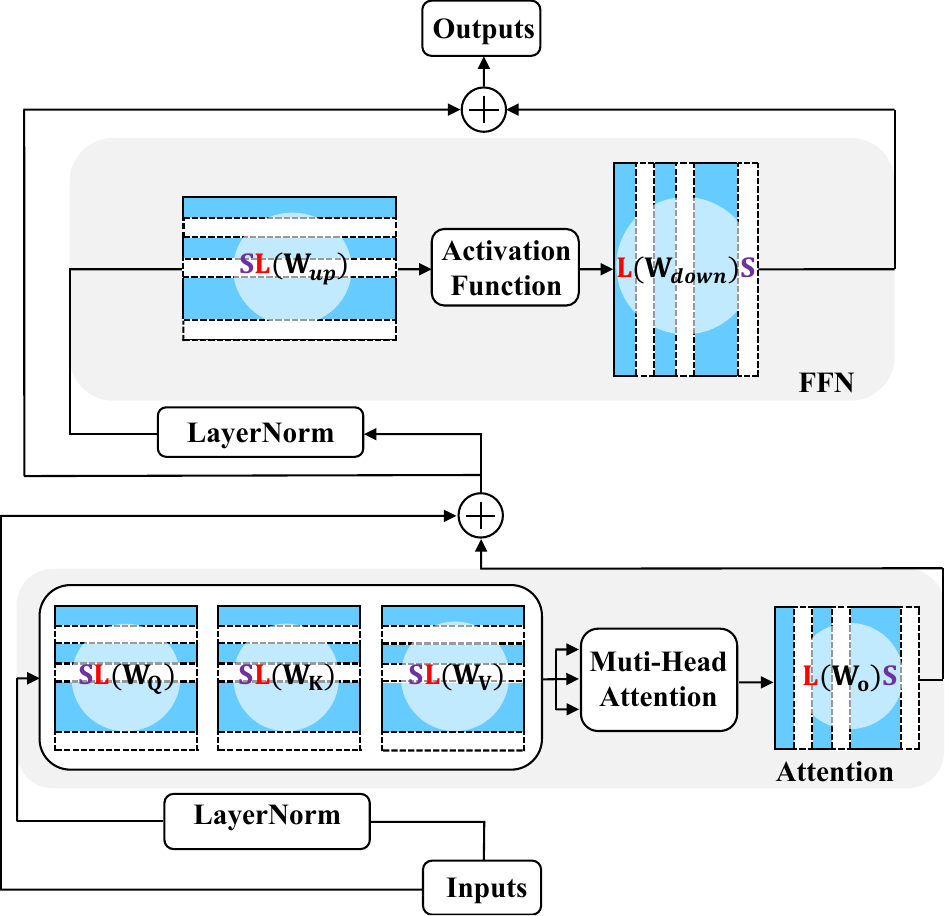}}
\caption{The regularization and pruning process of DReSS in a transformer layer.}
\label{picture2}
\end{center}
\end{figure}
\textbf{Proposition 1}
\label{proposition1}
\textnormal{(Dependency between attention and FFN block. Proof in Appendix A)}. If regularization is applied to certain columns of \( \mathbf{W_{down}^{i-1}} \) in the $(i-1)$th layer FFN block, then regularization should also be applied to the corresponding rows of \( \mathbf{W_{Q}^i} \), \( \mathbf{W_{K}^i} \), and \( \mathbf{W_{V}^i} \) in the $i$th layer attention block. The same dependency applies to \( \mathbf{W_{o}^i} \) and \( \mathbf{W_{up}^{i-1}} \).

Figure \ref{picture2} illustrates the process of applying regularization in a transformer layer. Following Proposition 1, in the attention block, regularization is applied to the rows of \( \mathbf{W_Q^i} \), \( \mathbf{W_K^i} \), and \( \mathbf{W_V^i} \) corresponding to the one entries in the index matrix \( \mathbf{R} \) and the columns of \( \mathbf{W_o^i} \) associated with these entries. In the FFN block, regularization is applied on the rows of \( \mathbf{W_{up}^i} \) and the columns of \( \mathbf{W_{down}^i} \) that correspond to the one entries in \( \mathbf{R} \). We also apply regularization to the remaining parts in LLMs. Specifically, for \(\mathbf{ W_{emb}} \), regularization is enforced on the columns corresponding to the one indices in \( \mathbf{R} \), and the same for \( \mathbf{W_{pos}} \). For each $\mathbf{\alpha^i}$ and $\mathbf{\beta^i}$ within every transformer layer, regularization is applied to the elements corresponding to the one indices in \( \mathbf{R} \). Additionally, for the final \( \mathbf{W_{lm}} \), due to its dependency on \( \mathbf{W_{down}^{l}} \) in the FFN block of the last transformer layer, regularization is applied to the corresponding rows. Therefore, the overall loss function defined in Equation~\ref{equation1} consists of the language modeling loss $\mathcal{L}(\mathbf{W}, \mathbf{X})$ and the regularization loss. \(\mathbf{X}\) is the input data. Regularization loss has three parts: Attention loss: $\sum_{i=1}^{l} \tilde{\mathcal{L}}_{\text{att}}^i$, FFN loss: $\sum_{i=1}^{l} \tilde{\mathcal{L}}_{\text{FFN}}^i$, remain loss: $\tilde{\mathcal{L}}_{\text{remain}}$.
\begin{equation}
\label{equation1}
\begin{split}
\mathcal{L}_{\text{sum}} = \mathcal{L}(\mathbf{W}, \mathbf{X}) +\sum_{i=1}^{l} \tilde{\mathcal{L}}_{\text{att}}^i + \sum_{i=1}^{l} \tilde{\mathcal{L}}_{\text{FFN}}^i + 
\tilde{\mathcal{L}}_{\text{remain}}
\end{split}
\end{equation}
$\tilde{\mathcal{L}}_{\text{att}}^i$ denotes the $i$th layer attention block regularization loss: 
\begin{equation}
\label{equation2}
\begin{split}
\tilde{\mathcal{L}}_{\text{att}}^i=\lambda(\|\mathbf{RW_Q^i}\|+\|\mathbf{RW_K^i}\|+ \|\mathbf{RW_V^i}\|+\|\mathbf{W_o^i R}\|)
\end{split}
\end{equation}
$\tilde{\mathcal{L}}_{\text{FFN}}^i$ is the $i$th layer FFN block regularization loss:
\begin{equation}
\label{equation3}
\begin{split}
\tilde{\mathcal{L}}_{\text{FFN}}^i=\lambda\|\mathbf{RW_{up}^i}\| + \lambda\|\mathbf{W_{down}^i R}\|
\end{split}
\end{equation}
$\tilde{\mathcal{L}}_{\text{remain}}$ denotes the regularization loss of the remaining parts of the LLM:
\begin{equation}
\label{equation4}
\begin{split}
\tilde{\mathcal{L}}_{\text{remain}}=\lambda(\|\mathbf{W_{emb}R}\| +\|\mathbf{W_{pos}R}\|+\\\sum_{i=1}^{l}\|\mathbf{R\alpha^i}\|+\sum_{i=1}^{l}\|\mathbf{R\beta^i}\|+\|\mathbf{RW_{lm}}\|)
\end{split}
\end{equation}
In Equations \ref{equation2}, \ref{equation3}, and \ref{equation4}, $\lambda$ is the same, $\|\cdot\|$ is a certain vector norm applied to the matrix column-wise.

\textbf{Proposition 2}
\label{l1-norm}
\textnormal{(If the loss function includes the $\ell_1$-norm, it can also be solved using backpropagation. Proof in Appendix B)}. The following two problems are equivalent, where $\|\cdot\|_1$ denotes the \( \ell_1 \)-norm.
\begin{equation}
\label{equation5}
\begin{aligned}
\min \quad||x||_1  \quad \iff \quad \min_{x, y}& \quad\mathbf{1}^T y \\
\text{s.t.}& \quad -y \leq x \leq y, \\
\quad &\quad y \geq 0.
\end{aligned}
\end{equation}
We formalize the minimization of the overall loss as an optimization problem. When $\ell_2$-norm is used, the problem can be directly solved using backpropagation (BP) algorithm~\cite{rumelhart1986learning}. In contrast, when the $\ell_1$-norm is employed, as the regularization losses introduced by various components in LLMs are similar, we consider only $\tilde{\mathcal{L}}_{\text{FFN}}^i$ as an example, the complete objective function transformation is in Appendix C. According to Proposition 2, the optimization problem in Equation~\ref{equation6} can be equivalently transformed into a constrained formulation in Equation \ref{equation7}, enabling its solution by BP algorithm. 

During forward and backward propagation, regularization on certain channels in LLMs significantly reduces the values in those rows or columns. Intuitively, this value reduction could decrease the impact of these parameters on the model's performance thus potentially transfer important information to the unregularized parts.
\begin{equation}
\label{equation6}
\begin{aligned}
\min_{\mathbf{W}} \hspace{0.5em}\mathcal{L}(\mathbf{W}, \mathbf{X}) +\lambda||\mathbf{RW_{up}^i}||_1 +\lambda||\mathbf{W_{down}^i R}||_1 
\end{aligned}
\end{equation}
\begin{equation}
\label{equation7}
\begin{aligned}
\min_{\mathbf{W, Y_1,Y_2}}& \quad\mathcal{L}(\mathbf{W}, \mathbf{X}) + \lambda(\mathbf{1}^T \mathbf{Y_1} \mathbf{1} + \mathbf{1}^T \mathbf{Y_2} \mathbf{1}) \\
\text{s.t.}& \quad -\mathbf{Y_1} \leq \mathbf{RW_{up}^i} \leq \mathbf{Y_1}, \\
 &  \quad -\mathbf{Y_2} \leq \mathbf{W_{down}^i R} \leq \mathbf{Y_2},\\
  \quad& \quad \mathbf{Y_1} \geq 0,\quad \mathbf{Y_2} \geq 0.
\end{aligned}
\end{equation}
\subsection{Pruning}
According to Figure \ref{picture2}, we prune the regularized rows and columns using the pseudo-index selection matrix \( \mathbf{R} \). For clarity, we define \( \mathbf{S = I - R} \). The pruning of the \( i \)th layer attention block and FFN block are performed as: 
\begin{equation}
\label{equation8}
\begin{aligned}
\mathbf{{W_Q^i}'} &\mathbf{= S W_Q^i,} \quad & \mathbf{{W_K^i}'} &\mathbf{= S W_K^i,} \\
\mathbf{{W_V^i}'} &\mathbf{= S W_V^i,} \quad & \mathbf{{W_o^i}'} &\mathbf{= W_o^i S,}\\
\mathbf{{W_{up}^i}'} &\mathbf{= SW_{up}^i,} \quad & \mathbf{{W_{down}^i}'} &\mathbf{= W_{down}^i S}
\end{aligned}
\end{equation}
The remaining parts are pruned as below:
 \begin{equation}
\label{equation10}
\begin{aligned}
\mathbf{{W_{emb}}'= W_{emb}S, \quad {W_{pos}}' = W_{pos}S,}\\
\mathbf{{\alpha^i}'=S{\alpha^i},\quad{\beta^i}' =S{\beta^i},\quad{W_{lm}}' = SW_{lm}}
\end{aligned}
\end{equation}
The pruning process is straightforward and simple.
\subsection{Optional RFT}
After pruning, we perform RFT using a small subset of the data selected in Section~\ref{3.1}, leveraging LoRA~\cite{hu2021lora}. This step is optional, as experimental results in section \ref{4.9} show that DReSS without RFT still achieves competitive performance in perplexity and zero-shot accuracy.
\section{Experiments}
\begin{table*}[t]
\setlength{\tabcolsep}{1mm}
\centering
\begin{tabular}{cccccccccccc}
\toprule
\textbf{Model} & \textbf{Method} & \textbf{PR} & \textbf{PPL ($\downarrow$)} & \textbf{PIQA} & \textbf{WinoGrande} & \textbf{HellaSwag} & \textbf{ARC-e} & \textbf{ARC-c} & \textbf{Avg\_Acc} \\
\midrule
\multirow{6}{*}{\textbf{Phi-2}} 
 & Dense       & 0\%  &  5.28 & 79.11 & 75.77 & 73.83 & 78.32 & 54.18 & 72.24\\
 & LLM Surgeon    & 25\%  & 7.26 & 67.28 & 63.25 &  54.24&  51.62& 35.85 &54.44  \\
 & SliceGPT    & 25\%  &7.06  & \textbf{69.32} &\underline{65.39} & 52.57 &\textbf{53.78}  &31.89  &54.59  \\
 & SLEB & 25\%  & 7.82 &67.94  &62.79  & 49.80 & 48.55 & 29.34 &51.68  \\
 & DReSS-\(\ell_2\) & 25\% & \textbf{6.25} & 68.52 & \textbf{66.71} & \underline{56.73} & \underline{52.78} & \textbf{37.63} & \textbf{56.47} \\
 & DReSS-\(\ell_1\) & 25\% & \underline{6.28} & \underline{68.67} & 65.32 & \textbf{57.16} & 52.39 & \underline{37.28} & \underline{56.16} \\
\midrule
\multirow{6}{*}{\textbf{LLaMA2-7B}} 
 & Dense        & 0\%  & 5.47 & 79.11 &69.06 & 75.99 & 74.58 & 46.25 & 69.00 \\
 & LLM Surgeon    & 25\% & 7.38 & 70.59 & \underline{65.87} & 58.66 & 63.65 & 38.33 & 59.42 \\
 & SliceGPT    & 25\% & 7.49 & 68.15 & 64.13 & 56.22 & 55.39 & 34.74 & 55.73 \\
 & SLEB    & 25\% & 10.24 & 63.25 & 62.36 & 53.77 & 55.82 &  32.24& 53.49 \\
 & DReSS-\(\ell_2\) & 25\% & \underline{5.86} & \underline{73.18} & \textbf{66.49} & \underline{61.73} & \textbf{65.42} & \textbf{40.86} & \textbf{61.54} \\
 & DReSS-\(\ell_1\) & 25\% & \textbf{5.81} & \textbf{73.42} & 65.73 & \textbf{61.92} & \underline{65.26} & \underline{39.68} & \underline{61.20} \\
\midrule
\multirow{6}{*}{\textbf{LLaMA3-8B}} 
 & Dense  & 0\%  & 5.76  & 85.56 & 77.94 & 79.27 & 78.84 &56.49 & 75.62 \\
 & LLM Surgeon   & 25\% & 7.62  & 76.34 & 70.18 & 71.46 & 69.65 &49.22 & 67.37 \\
 & SliceGPT    & 25\%&8.14  &74.37 & 67.81 & 69.56 & 69.83 &45.53 &65.42 \\
 & SLEB    & 25\% & 11.37  & 71.68 & 62.96 & 66.37 & 64.61 &43.28 &61.78 \\
 & DReSS-\(\ell_2\) & 25\% & \textbf{6.09} & \underline{78.29} & \textbf{71.17} & \underline{73.28} & \textbf{72.64} & \textbf{53.62} & \textbf{69.80} \\
 & DReSS-\(\ell_1\) & 25\% & \underline{6.12} & \textbf{78.86} & \underline{70.28} & \textbf{73.85} & \underline{72.37} & \underline{53.02} & \underline{69.68} \\
\midrule
\multirow{6}{*}{\textbf{OPT-13B}} 
 & Dense       & 0\%  &10.12 & 76.82 & 64.80& 69.81 & 61.87 & 35.67 &61.79 \\
 & LLM Surgeon    & 25\% & 11.02 & \textbf{74.18} & 64.33 &  65.37 & 60.96 & \underline{34.98} & 59.96 \\
 & SliceGPT    & 25\% & 10.90 &73.72 & 64.28 & 63.33 &60.59 & 34.66 & 59.32 \\
 & SLEB    & 25\% & 12.02 & 72.62 & 63.96 & 62.79 & 59.21 & 34.12 & 58.50 \\
 & DReSS-\(\ell_2\) & 25\% & \textbf{10.38} & \underline{74.06} & \textbf{64.57} & \textbf{66.82} & \textbf{61.56} & \textbf{35.33} & \textbf{60.47} \\
 & DReSS-\(\ell_1\) & 25\% & \underline{10.56} & 73.65 & \underline{64.38} & \underline{66.46} & \underline{61.15} & 34.74 & \underline{60.07} \\
\midrule
\multirow{6}{*}{\textbf{LLaMA2-13B}} 
 & Dense        & 0\%  & 4.88 & 80.47 & 72.22 & 79.39 &77.48 &49.23 & 71.76 \\
 &LLM Surgeon    & 25\% & 5.75 & \textbf{77.75} & 69.62 & 74.31 & 72.83 & 43.52 & 67.61 \\
 & SliceGPT    & 25\% & 6.16 & 69.69 & 68.45 &63.73 & 63.46 & 39.90 & 61.05 \\
 & SLEB    & 25\% & 7.39 & 66.93 & 65.87 & 55.48 & 60.06 & 35.14 & 56.70 \\
 & DReSS-\(\ell_2\) & 25\% & \textbf{5.12} & 76.14 & \textbf{71.22} & \underline{76.51} & \underline{73.45} & \textbf{46.87} & \textbf{68.84} \\
 & DReSS-\(\ell_1\) & 25\% & \underline{5.16} & \underline{77.24} & \underline{70.61} & \textbf{76.75} & \textbf{73.84} & \underline{45.46} & \underline{68.78} \\
\bottomrule
\end{tabular}
\caption{Performance comparison of different pruning methods. `PPL' refers to the perplexity on WikiText-2. The accuracy (\%) is reported on five zero-shot benchmarks. The best result is highlighted in \textbf{bold}, and the second-best is \underline{underlined}. DReSS-\(\ell_2\) denotes the use of the \(\ell_2\)-norm, while DReSS-\(\ell_1\) denotes the use of the \(\ell_1\)-norm. The results below are all obtained after RFT.}
\label{table1}
\end{table*}
\subsection{Experimental Setup}
\label{4.1}
\textbf{Implementation:} All methods are implemented in PyTorch~\cite{paszke2019pytorch}, using the Hugging Face Transformers library~\cite{wolf2019huggingface}. All experiments are conducted on 80GB NVIDIA A100 GPUs. For fairness, we use llm-eval-harness~\cite{eval-harness} to evaluate. 


\textbf{Datasets:} For generation task, we evaluate the model’s perplexity on WikiText-2~\cite{ashkboos2024slicegpt}. For zero-shot task, the benchmarks are PIQA~\cite{bisk2020piqa}, WinoGrande~\cite{sakaguchi2021winogrande}, HellaSwag~\cite{zellers2019hellaswag}, ARC-e and ARC-c~\cite{clark2018think}. In Appendix H we use data from Alpaca~\cite{alpaca}, WikiText-2~\cite{merity2016pointer}, PTB~\cite{marcus1993building}, and C4~\cite{raffel2020exploring}. 

\textbf{Models:} We use models commonly adopted in the pruning domain including the LLaMA models (LLaMA2-7B, LLaMA3-8B, LLaMA2-13B)~\cite{touvron2023LLaMA,grattafiori2024llama}, OPT model (OPT-13B)~\cite{zhang2022opt}, and Phi-2~\cite{javaheripi2023phi}.

\textbf{Baselines:} We evaluate DReSS against competitive structured pruning methods: LLM Surgeon~\cite{van2023llm}, SliceGPT~\cite{ashkboos2024slicegpt}, and SLEB~\cite{song2024sleb}. 

\textbf{Evaluation Metrics:} The performance on generation task is measured by \textit{perplexity}\cite{yao2022zeroquant}, while zero-shot tasks performance is evaluated using \textit{accuracy}~\cite{dong2024pruner}. The acceleration effects are represented by \textit{throughput} and \textit{latency}~\cite{song2024sleb}.

\begin{table*}[t]
\centering
    \begin{tabular}{ccccccccccccc}
    \toprule
    \textbf{Model} & \textbf{Method} & \textbf{PR} & \textbf{PPL($\downarrow$)} & \textbf{Tokens/s($\uparrow$)} & \textbf{TI($\uparrow$)} & \textbf{Latency($\downarrow$)} & \textbf{Speedup($\uparrow$)} \\
    \midrule
    \multirow{3}{*}{\textbf{OPT-13B}} 
    & Dense       & 0\% & 10.12 & 1029 & 1.00× & 386.5 & 1.00× \\
    & DReSS & 25\% & 10.38 & 1194 & 1.16× & 319.42 & 1.21× \\
    & DReSS & 50\% & 13.85 & 1389 & 1.35× & 274.11 & 1.41× \\
    \midrule
    \multirow{3}{*}{\textbf{LLaMA2-13B}} 
    & Dense        & 0\% & 4.88 & 1066 & 1.00× & 396.9 & 1.00× \\
    & DReSS & 25\% & 5.12 & 1215 & 1.14× & 330.8 & 1.20× \\
    & DReSS & 50\% & 7.59 & 1407 & 1.32× & 285.5 & 1.39× \\
    \bottomrule
    \end{tabular}
    \caption{Comparison of throughput and latency under different ratios on OPT-13B and LLaMA2-13B. `PPL’ refers to the perplexity on Wikitext2, `PR' represents `pruning ratio', `TI' represents `throughput increase'.}
\label{table2}
\end{table*}
\subsection{Performance Comparison}
\label{4.2}
For fairness, we randomly selected 1,000 samples from WikiText-2 training set with sequence length of 2048 for each method. We ensure that the data used for regularization in DReSS is consistent with the calibration data used by LLM Surgeon, SliceGPT, and SLEB, and the data used for RFT in each method is also the same. The ratio of calibartion and RFT data was set to 3:1, the reason is discussed in Section~\ref{4.9}. All methods are implemented under this setup. The pruning ratio was 25\%. The last 25\% of the values in the diagonal matrix \(\mathbf{R}\) are set to 1, which means pruning the last 25\% of the rows or columns of the parameter matrix and we discussed the choice of \(\mathbf{R}\) in Section~\ref{4.5}.

As shown in Table~\ref{table1}, DReSS achieves superior performance in both \textit{generation} and \textit{zero-shot} tasks. On LLaMA2-7B, its perplexity is 20\% lower than the second-best method, LLM Surgeon. On OPT-13B, compared to the dense model, DReSS only drops 1\% in average accuracy. The performance difference between \(\ell_2\)-norm and \(\ell_1\)-norm is tiny. In the following, DReSS refers to the \(\ell_2\)-norm version.
\subsection{Acceleration Effectiveness}
\label{4.4}
Language processing in LLMs comprises two primary stages: prompt processing, which is compute-bound, and token generation, which is memory-bound. We separately analyze the speedup achieved in each stage. Table~\ref{table2} presents the throughput and latency results for OPT-13B and LLaMA2-13B, evaluated using a single 80GB NVIDIA A100 GPU. Following the methodology of previous work~\cite{song2024sleb}, the token generation test involves generating sentences of 128 tokens with a batch size of 64, whereas for prompt processing, latency is measured by processing an input sequence of 2048 tokens.

At a pruning ratio of 50\% on OPT-13B, DReSS delivers a 35\% improvement in throughput and a 30\% reduction in latency compared to the dense model. These results highlight the superior efficiency of DReSS in acceleration.
\begin{table}[ht]
    \centering
\begin{tabular}{cccc}
    \toprule 
    \textbf{Cases} & (1) & (2)  & (3) \\
    \midrule
    \textbf{PPL($\downarrow$)} & 5.86 & 5.89 & 5.87 \\
    \midrule
    \textbf{Avg\_Acc(\%)} & 61.54 & 61.39 & 61.46 \\
    \bottomrule
    \end{tabular}
        \caption{The impact of pruning different channels on LLaMA2-7B. `PPL' is the perplexity on WikiText-2. Avg\_Acc is on five zero-shot benchmarks.}
    \label{table3}
\end{table}
\subsection{The Impact of Pruning Different Channels}
\label{4.5}
Matrix \(\mathbf{R}\) is used to select the channels on which regularization is applied (followed by pruning). We choose three options of \(\mathbf{R}\) under 25\% sparsity: (1) remove the matrix's last 25\% rows or columns, (2) remove the first 25\% rows or columns , (3) divide the matrix into 5 segments, removing each segment's 5\% of the last rows or columns. The results on LLaMA2-7B are shown in Table~\ref{table3}. The performance differences among (1), (2), and (3) are minimal, demonstrating that DReSS is insensitive to pruning different channels.
\subsection{Robustness to Different Pruning Ratios}
\label{4.6}
\begin{figure}[ht]
\begin{center}
\centerline{\includegraphics[width=\columnwidth]{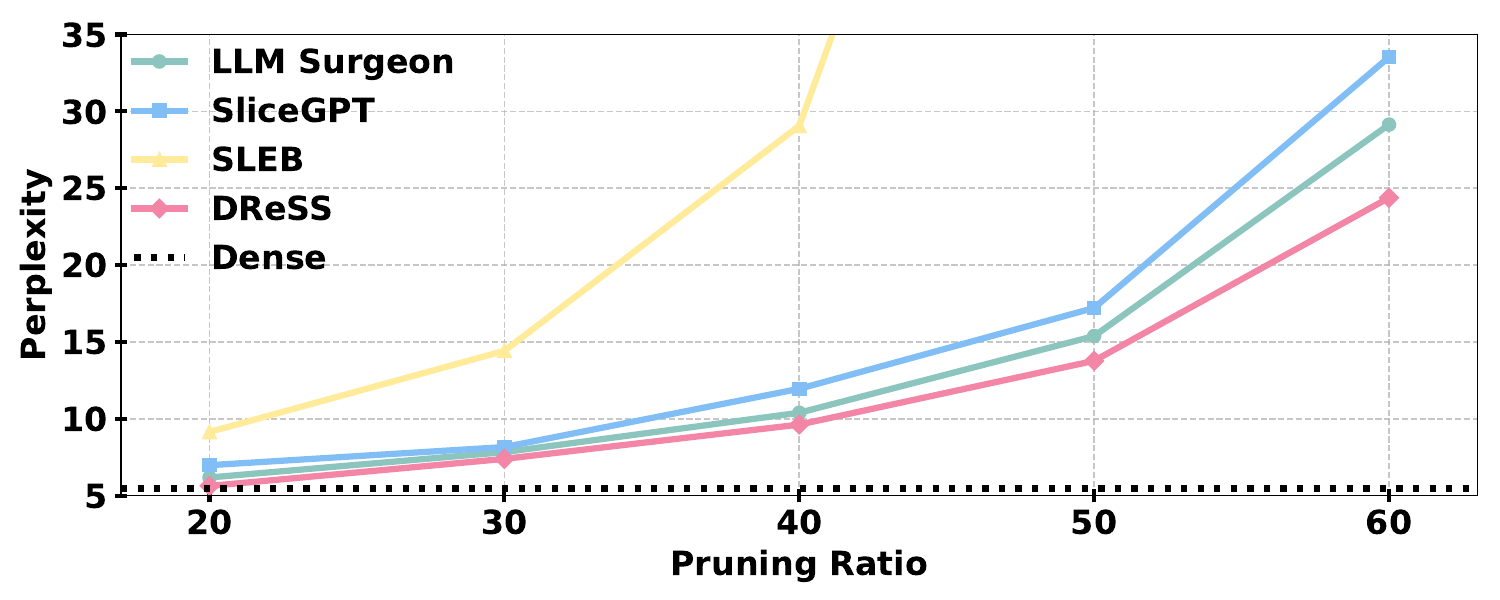}}
\caption{Perplexity of LLaMA2-7B pruned by various approaches under different pruning ratios on WikiText-2.}
\label{picture3}
\end{center}
\end{figure}

Keeping all other settings consistent with Section \ref{4.2}, we extend the pruning ratio from 20\% to 60\%. The perplexity of LLaMA2-7B under different methods are shown in Figure \ref{picture3}. DReSS significantly outperforms other methods across various pruning ratios. When the pruning ratio is up to 60\%, SLEB collapses, while DReSS maintains relatively low perplexity compared to other pruning methods. This demonstrates that DReSS maintains robust performance even under extreme pruning ratios, enabling structured pruning to unlock significant potential for model acceleration. More detailed results are in Appendix G.
\begin{figure}[ht]
\begin{center}
\centerline{\includegraphics[width=\columnwidth]{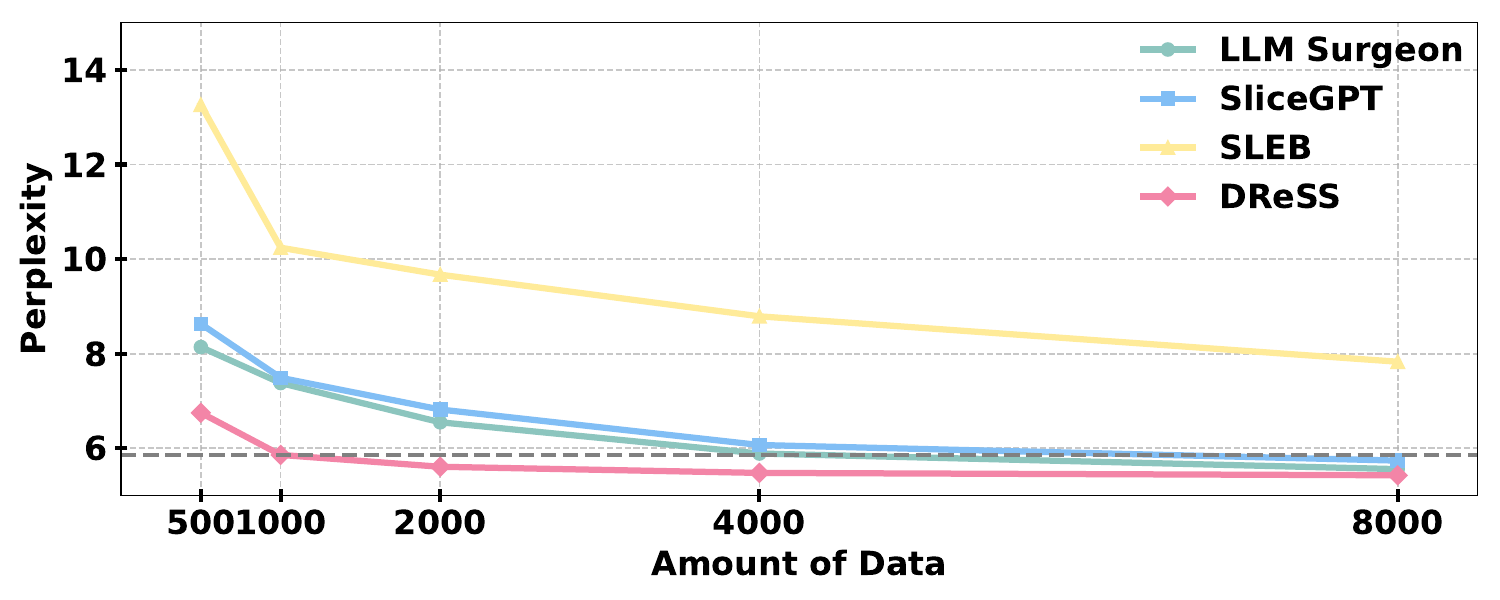}}
\caption{Perplexity of DReSS and other methods on WikiText-2 under different amount of data.}
\label{picture5}
\end{center}
\end{figure}
\begin{figure*}[ht]
\centering
\includegraphics[width=\textwidth]{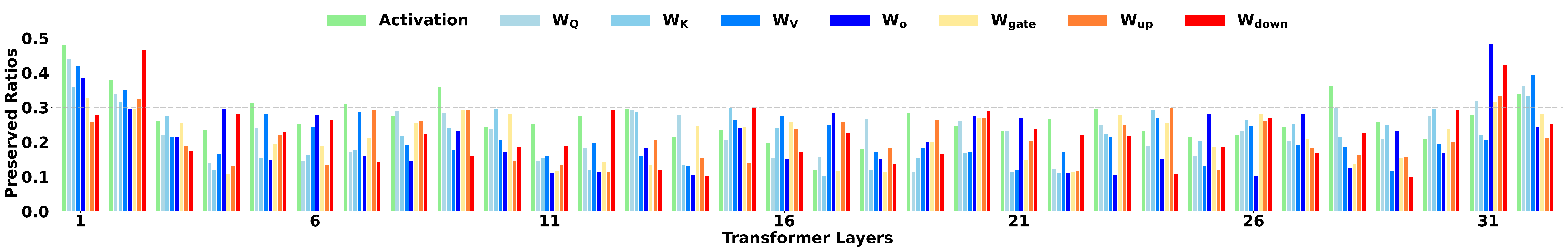}
\caption{The ratio of the sum of absolute parameter values after regularization to that before regularization on LLaMA2-7B. For example, the absolute sum of the last 25\% of rows in $\mathbf{W_Q}$ after regularization is divided by the sum of the corresponding rows before regularization to obtain the preserved ratios. This is applied similarly to other matrices $\mathbf{W_K, W_V, W_o, W_{gate}, W_{up},W_{down}}$, and layer activations.}
\label{picture6}
\end{figure*}
\begin{table*}[ht]
\centering
\begin{tabular}{ccccccccc}
\toprule 
\textbf{Cases} &N\&N\&N & R\&N\&N & R\&N\&R & N\&N\&R & R\&P\&N& R\&P\&R & N\&P\&N & N\&P\&R\\
\midrule
\textbf{PPL($\downarrow$)} &5.47& 5.06 & 4.86 & 5.06 &5.97 & 5.63&22.38&11.45\\
\midrule
\textbf{Avg\_Acc(\%)} &69.00& 70.39 & 71.23 & 70.39 &59.56&62.35&47.25&52.73 \\
\bottomrule
\end{tabular}
\caption{The model performance under different regularization settings. ‘PPL’ is the perplexity on WikiText-2. Avg\_Acc is on five zero-shot benchmarks. Pruning ration is 25\% and the LLM is LLaMA2-7B. We didn't perform RFT after pruning.}
\label{table7}
\end{table*}
\subsection{Minimal Overhead}
\label{4.8}
We evaluated the perplexity of each method on LLaMA2-7B using varying amounts of data, keeping all other conditions consistent with Section~\ref{4.2} and only change the data size from 500 to 8,000. As shown in Figure~\ref{picture5}, when only 1,000 samples were used for regularization and RFT, DReSS outperformed the other methods that used 4,000 samples. This further highlights the effectiveness and efficiency of applying regularization prior to pruning, as it transfers critical information in advance, enabling DReSS to get strong performance with minimal data overhead.
\subsection{Ablation Study}
\label{4.9}
As shown in Table~\ref{table4}, both pruning without regularization and full parameter regularization significantly increase perplexity on WikiText-2 and reduce average accuracy at 25\% sparsity. This highlights the importance of applying regularization precisely to the pruned components. Additionally, RFT after pruning has minimal impact on model performance, indicating that the RFT process is optional.

\textbf{Data Ratio of Regualrization and RFT:} With other settings consistent with Section~\ref{4.2}, we only vary the data ratio \([4:1, 3:1, 2:1, 1:1, 1:2, 1:3]\). The lowest perplexity is achieved under 3:1 ratio. We speculate that pre-pruning regularization may be more critical than post-pruning RFT.
\begin{table}[ht]
    \centering
    \begin{tabular}{cccc}
    \toprule
    \textbf{Model} & \textbf{Setting} & \textbf{PPL($\downarrow$)}  & \textbf{AVG\_ACC}  \\
    \midrule
    \multirow{4}{*}{\textbf{LLaMA2-7B}} 
    &DReSS & 5.86  & 61.54  \\
    &w FPR & 12.68  & 50.26  \\
    &w/o R & 22.38  & 47.25  \\
    &w/o RFT & 5.97  & 59.56  \\
    \midrule
    \multirow{4}{*}{\textbf{LLaMA2-13B}} 
    &DReSS & 5.12 & 68.84  \\
    &w FPR & 10.53  & 53.28  \\
    &w/o R & 18.94  & 50.17  \\
    &w/o RFT & 5.39  & 67.96  \\
    \bottomrule
    \end{tabular}
        \caption{Ablation results on LLaMA2-7B, LLaMA2-13B. `FPR' is full parameter regularization, `R' is regularization on selected channels. `w/o' means removing specific parts.}
    \label{table4}
\end{table}

\textbf{Trade-Off Between Language Modeling and Regularization:} Keeping all other settings consistent with Section~\ref{4.2}, we evaluate the model's performance by varying \(\lambda \in [10^{-5}, 10^{-4}, 10^{-3}, 5 \times 10^{-3}, 10^{-2}, 5 \times 10^{-2}, 10^{-1}]\). Optimal $\lambda$ is $10^{-3}$, demonstrating the importance of balancing language modeling loss and regularization loss. We listed the best \(\lambda\) for each model in Appendix E.

\subsection{Impact of Regularization}
\label{4.3}
In Figure \ref{picture6}, we present the ratio of model weights and activation values after and before regularization of the selected portions. We find that in most intermediate layers, the weights and activation values after regularization are reduced to 30\% of their original values, which indicates that information contained in this parameter space has decreased. We also present the changes in the unregularized part in Appendix F. The results show that this part has increased compared to before, indicating that it now contains more information showing that regularization transfers important information from the pruned parts to the remaining parts.

To evaluate regularization effectiveness, we divided into three stages: whether to regularize before pruning, whether to prune, and whether to regularize after pruning, totally 8 cases. For example,``N\&P\&R" denotes no regularization, pruning, then regularized. We used 25\% sparsity LLaMA2-7B.``N\&N\&R" and ``R\&N\&N" are the same. The 750 (0.75$\times$1,000) samples used for regularization are consistent with Section~\ref{4.2}. As shown in Table~\ref{table7}, ``R\&P\&N" and ``N\&P\&R" outperform ``N\&P\&N", ``R\&P\&R" is the best, showing that regularization improves performance before, after pruning, and both of them. ``R\&P\&N" outperforms ``N\&P\&R", indicating that the paradigm of applying regularization before pruning may transfer important information to the remaining parts, thereby preserving model capacity.
\section{Conclusion}
\label{5}
In this paper, we propose a new pruning paradigm: apply regularization, prune, and finally RFT. Unlike previous paradigm that first prune and then apply RFT, DReSS transfers critical information from the pruned parameter space to the remaining parts during regularization, effectively mitigating the irreversible performance degradation caused by information loss. DReSS demonstrates superior performance in generation and zero-shot tasks, significantly outperforming existing methods. For instance, DReSS surpasses the powerful LLM Surgeon by 21\% in perplexity on LLaMA2-7B. On OPT-13B, under 25\% sparsity, the average accuracy drops by only 1\%, while achieving 1.21× speedup compared to the dense model. Moreover, DReSS requires only 25\% of the data to achieve comparable performance to previous methods, substantially reducing computational costs and the reliance on RFT. The new paradigm may provide insights for structured pruning in LLMs.

\bibliography{aaai2026}
\newpage
\appendix
\onecolumn
\section{Proof of the Proposition 1}
\label{A}
Let us first assume that \( \mathbf{A} \in \mathbb{R}^{m \times n} \) and \( \mathbf{B} \in \mathbb{R}^{n \times k} \). We can express \(\mathbf{ A} \) as \( \begin{bmatrix} \mathbf{a}_1 & \mathbf{a}_2 & \dots & \mathbf{a}_n \end{bmatrix} \), where \( \mathbf{a}_i \in \mathbb{R}^{m \times 1} \), and \( \mathbf{B} \) as \( \begin{bmatrix} \mathbf{b}_1^T \\ \mathbf{b}_2^T \\ \vdots \\ \mathbf{b}_n^T \end{bmatrix} \), where \( \mathbf{b}_i \in \mathbb{R}^{k \times 1} \). Then, we obtain \( \mathbf{C} \in \mathbb{R}^{m \times k} \) as shown in Equation~\ref{equation11}. It is important to note that the final summation in Equation~\ref{equation11} refers to the summation of \( n \) rank-1 matrices, not a scalar summation.
\begin{equation}
\label{equation11}
\begin{split}
\mathbf{C }= \mathbf{AB} = \begin{bmatrix} \mathbf{a}_1 & \mathbf{a}_2 & \dots & \mathbf{a}_n \end{bmatrix} \times \begin{bmatrix} \mathbf{b}_1^T \\ \mathbf{b}_2^T \\ \vdots \\ \mathbf{b}_n^T \end{bmatrix}=\sum_{i=1}^{n}\mathbf{a}_i \mathbf{b}_i^T
\end{split}
\end{equation}

\textbf{According to Equation~\ref{equation11}, To keep the elements of $\mathbf{a}_i \mathbf{b}_i^T$ small, both $\mathbf{a}_i$ and $\mathbf{b}_i^T$ must be small. So, if regularization is applied to $\mathbf{a}_i$ and it should also be applied to the corresponding row $\mathbf{b}_i^T$ in $\mathbf{B}$.}

Applying the above conclusion to the parameters of a transformer layer, if regularization is applied to certain columns of \( \mathbf{W_{down}^{i-1}} \) in the FFN layer of the \( (i-1) \)th transformer layer, then, since \( \mathbf{W_Q^i} \), \( \mathbf{W_K^i} \), and \( \mathbf{W_V^i} \) in the \( i \)th transformer layer are immediately multiplied by the result of \( \mathbf{W_{down}^{i-1}} \), regularization should also be applied to the corresponding rows in \( \mathbf{W_Q^i} \), \(\mathbf{ W_K^i} \), and \( \mathbf{W_V^i} \). Additionally, if regularization is applied to certain columns of \( \mathbf{W_o^i} \) in the \( i \)th transformer layer, the \( \mathbf{W_{up}^i} \) in the FFN block will immediately multiply with it, so regularization should also be applied to the corresponding rows in \( \mathbf{W_{up}^i} \). Thus, we have completed the proof of Proposition 1.

\section{Proof of Proposition 2}
\label{B}

\textbf{Step 1: Expressing \(\ell_1\)-norm Using Elements.} \\
The objective function in the unconstrained problem is the \( \ell_1 \)-norm of the vector \( x \), which is defined as:
\begin{equation}
||x||_1 = \sum_{i=1}^{n} |x_i|
\end{equation}

This function aims to minimize the sum of the absolute values of the components of \( x \).

\textbf{Step 2: Reformulating the Constrained Problem} \\
The constrained optimization problem introduces an auxiliary variable \( y \), where for each element \( i \):
\begin{equation}
x_i \geq -y_i \quad \text{and} \quad   x_i \leq y_i\quad 
\end{equation}
This implies that \( y_i \geq |x_i| \), meaning each element of \( y \) serves as an upper bound for the absolute value of the corresponding element in \( x \). Consequently, minimizing \(|x|_1\) is equivalent to minimizing the sum of the elements in \( y \). Thus, the objective function is defined as:
\begin{equation}
\mathbf{1}^T y
\end{equation}
Thus, minimizing \( \mathbf{1}^T y \) is equivalent to minimizing the sum of the absolute values of \( x \), which is the \( \ell_1 \)-norm of \( x \).

This transformation allows the optimization problem to be solved without directly involving the absolute value function, resulting in an equivalent constrained optimization problem that can be addressed via backpropagation. Thus, the proof of the Proposition 2 is complete.

\section{Completed Training Objective Function}
\label{C}
The final training objective has two parts: language modeling loss $\mathcal{L}(\mathbf{W}, \mathbf{X})$, regularization loss. Regularization loss has three parts: Attention loss, FFN loss, remain loss. In Equation \ref{equation7}, for clarity, we only list one FFN layer, and principles for the others are similar. By using Proposition 2, the problem can be equivalently transformed into a constrained and differentiable optimization problem, which can be directly solved using the BP algorithm. The final objective function and constraints are as follows:

\begin{equation}
\begin{aligned}
\mathcal{L}_{\text{sum}} &= \mathcal{L}(\mathbf{W, X}) 
+ \sum_{i=1}^{l} \tilde{\mathcal{L}}_{\text{att}}^i 
+ \sum_{i=1}^{l} \tilde{\mathcal{L}}_{\text{FFN}}^i 
+ \tilde{\mathcal{L}}_{\text{remain}} \\
\tilde{\mathcal{L}}_{\text{att}}^i &= \lambda(\mathbf{\mathbf{1}^T Y_3^i \mathbf{1} + \mathbf{1}^T Y_4^i \mathbf{1} + \mathbf{1}^T Y_5^i \mathbf{1} + \mathbf{1}^T Y_6^i \mathbf{1}}) \\
\tilde{\mathcal{L}}_{\text{FFN}}^i &= \lambda(\mathbf{\mathbf{1}^T Y_1^i \mathbf{1} + \mathbf{1}^T Y_2^i \mathbf{1}}) \\
\tilde{\mathcal{L}}_{\text{remain}} &= \lambda\Big(
\mathbf{1}^T \mathbf{Y_7}\mathbf{1} + \mathbf{1}^T \mathbf{Y_8} \mathbf{1} 
+ \sum_{i=1}^{l} \mathbf{1}^T y_9^i \mathbf{1}+\sum_{i=1}^{l} \mathbf{1}^T y_{10}^i \mathbf{1} + \mathbf{1}^T \mathbf{Y_{11}} \mathbf{1} \Big) \\
\text{s.t.}\quad
- \mathbf{Y_1^i} &\mathbf{\leq R W_{up}^i \leq Y_1^i},\quad
- \mathbf{Y_2^i \leq W_{down}^i R \leq Y_2^i,\quad
- Y_3^i \leq R W_Q^i \leq Y_3^i}, \\
- \mathbf{Y_4^i} &\mathbf{\leq R W_K^i \leq Y_4^i,\quad
- Y_5^i \leq R W_V^i \leq Y_5^i,\quad
- Y_6^i \leq W_o^i R \leq Y_6^i}, \\
- \mathbf{Y_7} &\mathbf{\leq W_{emb} R \leq Y_7,\quad
- Y_8 \leq W_{pos} R \leq Y_8},\quad
- y_9^i \mathbf{\leq R \alpha^i} \leq y_9^i, \\
- y_{10}^i& \mathbf{\leq R \beta^i} \leq y_{10}^i, \quad
- \mathbf{Y_{10} \leq R W_{lm} \leq Y_{10}} \\
\mathbf{Y_1^i} &\mathbf{\geq 0,\quad Y_2^i \geq 0,\quad Y_3^i \geq 0,\quad Y_4^i \geq 0,\quad Y_5^i \geq 0,\quad Y_6^i \geq 0}, \\
\mathbf{Y_7} &\mathbf{\geq 0,\quad Y_8 \geq 0},\quad y_9^i \geq 0,\quad y_{10}^i \geq 0,\quad \mathbf{Y_{11}} \geq 0
\end{aligned}
\end{equation}

\section{Detailed Implementation}
\label{D}
In this part, we first introduce several hyperparameter settings, with the detailed results shown in Table \ref{table9}. In our experiments, we employ FP16 precision for all evaluated models, including Phi-2, OPT-2.7B, LLaMA3-8B, OPT-13B, LLaMA2-7B, and LLaMA2-13B. For all RFT configurations, we set the LoRA rank \( r \) to 32, the scaling factor \( \alpha \) to 10, and the sequence length to 2048. All other hyperparameters follow the default settings provided in the Hugging Face PEFT package \cite{mangrulkar2022peft}. We set the batch size to 64. In future work, we will further explore a broader range of batch sizes.
\begin{table}[ht]
\centering
\begin{tabular}{cccccccccccc}
\toprule
 \textbf{LoRA Rank} & \textbf{Scaling Factor}& \textbf{Max Sequence Length} & \textbf{Batch Size} & \textbf{Learning Rate} & \textbf{Early Stop Threshold}   \\
\midrule 
 32       & 10 &2048 &64 & 2e-5 & 5  \\
\bottomrule
\end{tabular}
\caption{Implementation Details}
\label{table9}
\end{table}
To ensure a fair comparison between DReSS and other methods, we maintain consistency in the data used across all approaches. Specifically, the data used by DReSS for regularization, by LLM Surgeon for periodic updates of model weights and structures, by SliceGPT for selecting channel importance, and by SLEB for identifying crucial transformer layers are identical. Furthermore, we ensure that the data employed during the RFT process is consistent across all methods, thereby enabling a controlled and equitable evaluation framework. Following previous works~\cite{ashkboos2024slicegpt,song2024sleb}, for the comparison unstructured pruning methods like Mangnitude, Wanda, and SparseGPT, we ensure that the data used to compute the importance of individual weights is the same as the data used by DReSS for regularization.

\section{Optimal $\lambda$ for Different Models}
\label{E}
Keeping all other settings consistent with Section~\ref{4.2}, we evaluate the model's performance by varying \(\lambda \in [10^{-5}, 10^{-4}, 10^{-3}, 5 \times 10^{-3}, 10^{-2}, 5 \times 10^{-2}, 10^{-1}]\).

we list $\lambda$ for the best performance of each model in the Table~\ref{table10}:
\begin{table}[ht]
\centering

\begin{tabular}{ccccccc}
\toprule 
\textbf{Model} &Phi-2 & OPT-2.7B & OPT-6.7B & OPT-13B & LLaMA2-7B& LLaMA2-13B\\
\midrule
\textbf{$\lambda$} &$10^{-3}$& $5 \times 10^{-3}$ & $10^{-3}$ &$10^{-3}$ &$10^{-3}$ & $10^{-3}$\\
\bottomrule
\end{tabular}
\caption{The optimal \(\lambda\) for each model.}
\label{table10}
\end{table}

\section{Changes in the Parts Without Regularization}
\label{F}
\begin{figure}[!htbp]
\centering
\includegraphics[width=\textwidth]{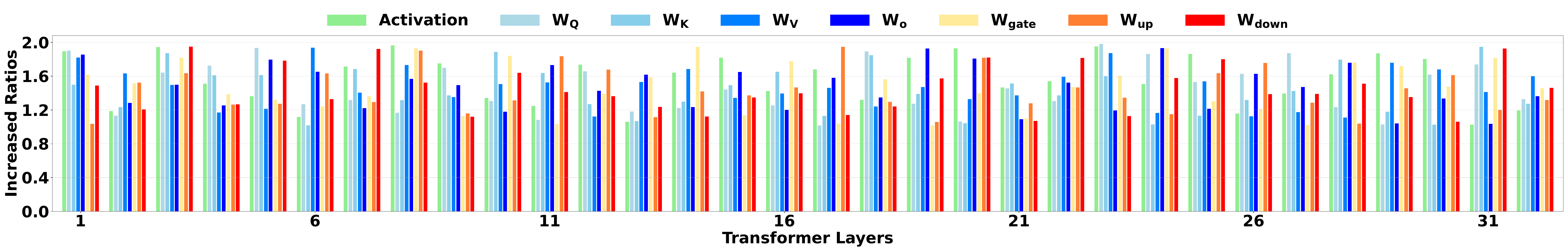}
\caption{Ratio of the sum of absolute values of unregularized parameters after LLaMA2-7B regularization to the sum of absolute values of the corresponding parameters before regularization. For example, regularization is not applied to the first 75\% of rows in $\mathbf{W_Q}$, and the absolute sum of the rows after regularization is divided by the sum of the corresponding rows before regularization to obtain the Increased Ratios. This is applied similarly to other matrices such as $\mathbf{W_K, W_V, W_o, W_{gate}, W_{up}, W_{down}}$, and for layer activations, the sum of the unregularized columns is compared to the original sum.}
\label{picture7}
\end{figure}

As illustrated in Figure \ref{picture7}, the magnitude of the unregularized parameters exhibits an increase after the application of regularization, suggesting a redistribution of model capacity. This phenomenon indicates that during regularization, critical information, initially encoded in the regularized portion of the model, is partially transferred to the unregularized portion. In contrast, Figure \ref{picture3} shows a reduction in the magnitude of the regularized parameters after regularization, implying that the imposed constraints effectively suppress the corresponding parameter values, enforcing sparsity or compression in that region. 

Collectively, these observations suggest that the regularization process facilitates an implicit redistribution of information across different parameter subsets. Specifically, the regularization term promotes a shift of important model characteristics from the constrained (regularized) portion to the unconstrained (unregularized) portion, thereby preserving essential model knowledge despite the imposed sparsity constraints. Based on this insight, we posit that pruning the regularized portion post-regularization could mitigate information loss, as the core knowledge has already been migrated to the unregularized segment. This pruning strategy effectively reduces parameter redundancy while retaining the model’s language modeling capacity, thereby achieving a more compact and efficient representation without compromising performance.

\section{Performance of DReSS under Different Pruning Ratios and Datasets}
\label{G}
\subsection{The Perplexity of DReSS under Different Pruning Ratios and Datasets}
In Section \ref{4.2}, we utilize 1,000 samples randomly selected from the WikiText-2 dataset to guide the regularization process. Subsequently, we evaluate multiple large language models (LLMs) by measuring changes in perplexity across various generative task datasets, including WikiText-2, Alpaca, PTB, and C4, under pruning rates of 10\%, 20\%, 30\%, 40\%, 50\%, and 60\%. The detailed results, presented in Table \ref{table12}, indicate that DReSS exhibits greater robustness as model scale increases, suggesting that the proposed method effectively mitigates performance degradation in larger architectures. This highlights the scalability of DReSS and its potential to maintain model efficiency under varying levels of sparsity.

\begin{table*}[!htbp]
\centering

\begin{tabular}{cccccc}
\toprule
\textbf{Model} & \textbf{Pruning Ratio}  & \textbf{WikiText-2} & \textbf{Alpaca} & \textbf{PTB} & \textbf{C4} \\
\midrule
\multirow{7}{*}{\textbf{OPT-2.7B}} 
 & Dense      & 12.46  &11.64  & 17.97 & 14.32 \\
 & 10\%       & 12.48  &11.71  & 18.16 & 14.71 \\
 & 20\%       & 12.50  &11.93  & 19.45 & 15.42 \\
 & 30\%       & 14.49  &12.88  & 23.58 & 18.93 \\ 
 & 40\%       & 18.92  &15.46  & 31.30 & 24.33 \\
 & 50\%       & 24.57  &21.37  & 45.32 & 32.15 \\
 & 60\%       & 33.83  &31.22  & 58.73 & 45.56 \\
\midrule
\multirow{7}{*}{\textbf{OPT-6.7B}} 
 & Dense      & 10.85  &10.27  & 15.77 & 12.71 \\
 & 10\%       & 10.91  &10.45  & 16.05 & 13.18 \\
 & 20\%       & 11.02  &10.83  & 17.58 & 14.45 \\
 & 30\%       & 12.32  &11.46  & 19.65 & 16.68 \\ 
 & 40\%       & 14.26  &12.71  & 25.52 & 20.94 \\
 & 50\%       & 19.63  &15.66  & 33.78 & 28.22 \\
 & 60\%       & 28.75  &21.39  & 47.29 & 39.47 \\
\midrule
\multirow{7}{*}{\textbf{OPT-13B}} 
 & Dense      & 10.12  &9.46  &14.52 & 12.06 \\
 & 10\%       & 10.22  &9.65  & 14.88 & 12.37 \\
 & 20\%       & 10.31  &9.97 & 15.64 & 13.15 \\
 & 30\%       & 10.99  &10.56  & 18.81 & 15.63 \\ 
 & 40\%       & 11.62 & 11.25  & 23.07 & 19.55 \\
 & 50\%       & 13.85  & 13.23  & 29.26 & 26.21 \\
 & 60\%       & 27.63  & 20.12 & 38.59 & 35.58 \\
 \midrule
\multirow{7}{*}{\textbf{LLaMA2-7B}} 
 & Dense      & 5.47  &5.25  & 7.92 & 7.26 \\
 & 10\%       & 5.54  &5.29  & 8.06 & 7.34 \\
 & 20\%       & 5.63  &5.37 & 8.29 &  7.79 \\
 & 30\%       & 7.39 & 7.32  & 9.13 &8.46 \\ 
 & 40\%       & 9.62  & 8.62 & 12.37 & 10.56 \\
 & 50\%       & 13.77  &12.59  & 18.85 & 15.18 \\
 & 60\%       & 24.38 &20.25 & 29.34 & 27.37 \\
\midrule
\multirow{7}{*}{\textbf{LLaMA2-13B}} 
 & Dense      & 4.88  &4.63  & 7.16 & 6.73 \\
 & 10\%       & 4.94  &4.69  & 7.23 & 6.96 \\
 & 20\%       & 5.08  &4.83  & 7.61 & 7.53 \\
 & 30\%       & 5.61  &5.42  & 8.33 & 8.14 \\ 
 & 40\%       & 6.25  &6.14  & 9.27 & 9.05 \\
 & 50\%       & 7.59  &7.28  & 11.58 & 10.88\\
 & 60\%       & 12.77 &11.78 & 15.16 & 13.62 \\
\bottomrule
\end{tabular}
\caption{Perplexity comparison of DReSS with different pruning ratios. We set the pruning ratios to 10\%, 20\%, 30\%, 40\%, 50\%, and 60\%, and test the perplexity of the OPT and LLaMA2 models on the generation task datasets Alpaca, WikiText-2, PTB, and C4. For DReSS, we use the $\ell_2$-norm.}
\label{table12}
\end{table*}

\subsection{The Accuracy of DReSS under Different Pruning Ratios on Zero-shot Tasks}
\label{G.2}
To systematically evaluate the performance of DReSS on zero-shot tasks under varying pruning rates, we adopt the experimental setup outlined in Section \ref{4.2}, where 1,000 samples are randomly selected from the WikiText-2 dataset to guide the regularization process. We assess the accuracy of different model configurations at pruning rates of 10\%, 20\%, 30\%, 40\%, 50\%, and 60\% across a diverse set of benchmark datasets, including PIQA, WinoGrande, HellaSwag, ARC-e, and ARC-c. The results, summarized in Table \ref{table13}, provide insights into the impact of sparsity on zero-shot generalization. Notably, the analysis reveals that DReSS maintains competitive performance even at higher pruning rates, demonstrating its effectiveness in preserving reasoning and commonsense understanding across different tasks.
\begin{table*}[!htbp]
\centering
\begin{tabular}{cccccccc}
\toprule
\textbf{Model} & \textbf{Pruning Ratio} & \textbf{PIQA} & \textbf{WinoGrande} & \textbf{HellaSwag} & \textbf{ARC-e}&\textbf{ARC-c }&\textbf{Avg\_Acc} \\
\midrule
\multirow{7}{*}{\textbf{OPT-2.7B}} 
 & Dense      & 74.81   &61.01 &60.58& 54.42&31.14 &56.39  \\
 & 10\%        &70.38  &60.12  &51.79 & 52.46&29.78 & 52.91 \\
 & 20\%        & 69.96  &59.47  &50.45& 51.87& 28.62& 52.07 \\
 & 30\%       & 64.52  &58.76  &48.25 & 50.23&27.52 &49.86  \\ 
 & 40\%       & 61.32  &56.27 &47.39 & 48.26& 25.57&47.76 \\
 & 50\%        & 59.38  &53.46  &44.63 & 46.19&21.66 & 45.06 \\
 & 60\%        & 52.99  &48.74  &40.03 &43.35&16.25 &40.27  \\
\midrule
\multirow{7}{*}{\textbf{OPT-6.7B}} 
 & Dense      & 76.39  &65.19  &67.16& 60.14& 34.64 & 60.70 \\
 & 10\%        & 75.89  &64.61  &64.69 & 58.53& 33.47& 59.44 \\
 & 20\%        & 75.12  &64.23  &62.56 & 57.34& 32.95&58.44  \\
 & 30\%       & 72.52  &62.63  &58.27 & 54.48&29.99 &55.58  \\ 
 & 40\%       & 67.37  &58.59  &52.12 & 49.38& 26.87&50.87 \\
 & 50\%        & 61.48  &55.62  &46.46 & 47.68&22.97 & 46.84 \\
 & 60\%        & 55.73  &51.52  &43.38 & 45.85&18.62 & 43.02 \\
\midrule
\multirow{6}{*}{\textbf{OPT-13B}} 
 & Dense      & 76.82  &64.80  &69.81 &61.87&35.67 & 61.79 \\
 & 10\%        & 75.46  &64.69  &68.56 & 61.79&35.58 & 61.22 \\
 & 20\%        & 74.78  &64.61  &67.25 & 61.66& 35.43& 60.75 \\
 & 30\%       & 72.62  &63.26  &65.69 & 59.64& 32.57&58.76  \\ 
 & 40\%       & 68.67  &61.49  &62.74 & 55.98&29.26 & 55.63\\
 & 50\%        & 62.19  &56.46  &58.55 & 52.15&24.53 & 50.78 \\
 & 60\%        & 57.43  &52.72  &51.23 & 47.62& 21.05& 46.01 \\
 \midrule
\multirow{7}{*}{\textbf{LLaMA2-7B}} 
 & Dense      &79.11  &69.06 &75.99  & 74.58& 46.25&69.00  \\
 & 10\%        & 77.38 &68.16  &71.28 &69.26 &43.73 & 65.96 \\
 & 20\%        & 76.42  &67.35  &68.26 &66.73 &41.68 & 64.09 \\
 & 30\%       & 72.29  &64.87  &58.53 &63.48 &39.27 & 59.69 \\ 
 & 40\%       & 68.66  &61.49  &55.42 &58.61 &36.52 & 56.14\\
 & 50\%        & 61.32  &55.68  &51.79 & 52.35&31.55 & 50.54 \\
 & 60\%        & 56.45  &51.79  &48.96 &48.98 &27.26 &  46.69\\
\midrule
\multirow{7}{*}{\textbf{LLaMA2-13B}} 
 & Dense       & 80.47  &72.22  &79.39 & 77.48&49.23&71.76  \\
 & 10\%        & 79.35  &72.15  &77.42 & 76.92& 48.57& 70.88 \\
 & 20\%        & 78.27 &71.98   &76.89 & 75.43&47.62 & 70.04 \\
 & 30\%        & 75.49  &69.73  &73.54 & 72.87&45.41 & 67.41 \\ 
 & 40\%        & 73.56 &64.46   &67.75 & 68.45& 41.28& 63.10\\
 & 50\%        & 68.73 &59.82   &62.48 & 60.12&36.14 & 57.46 \\
 & 60\%        & 62.25 & 56.27  &59.93 & 53.37&29.77 & 52.32 \\
\bottomrule
\end{tabular}
\caption{Accuracy comparison of DReSS with different pruning ratios. We set the pruning ratios to 10\%, 20\%, 30\%, 40\%, 50\%, and 60\%, and test the accuracy of the OPT and LLaMA2 models on the zero-shot task datasets PIQA, WinoGrande, HellaSwag, ARC-e and ARC-c. For DReSS, we use the $\ell_2$-norm. `Avg\_Acc' represents the average accuracy.}
\label{table13}
\end{table*}

\section{Dependency on Calibration Dataset}
\label{H}
Since DReSS relies on data-driven regularization, we investigate its dataset dependency. We evaluated perplexity of four methods on WikiText-2, using calibration and RFT data selected from Alpaca, WikiText-2, PTB, and C4. For fairness, we randomly selected 1,000 samples from each dataset, with other settings consistent with Section~\ref{4.2}. As shown in Figure~\ref{picture4}, DReSS consistently outperforms the other methods across datasets, demonstrating its robustness. When using Alpaca, WikiText-2, C4, and PTB as calibration and RFT data, the perplexity of various methods on WikiText-2, Alpaca, C4, and PTB is shown as follows:
\begin{figure}[ht]
\begin{center}
\centerline{\includegraphics[width=0.6\textwidth]{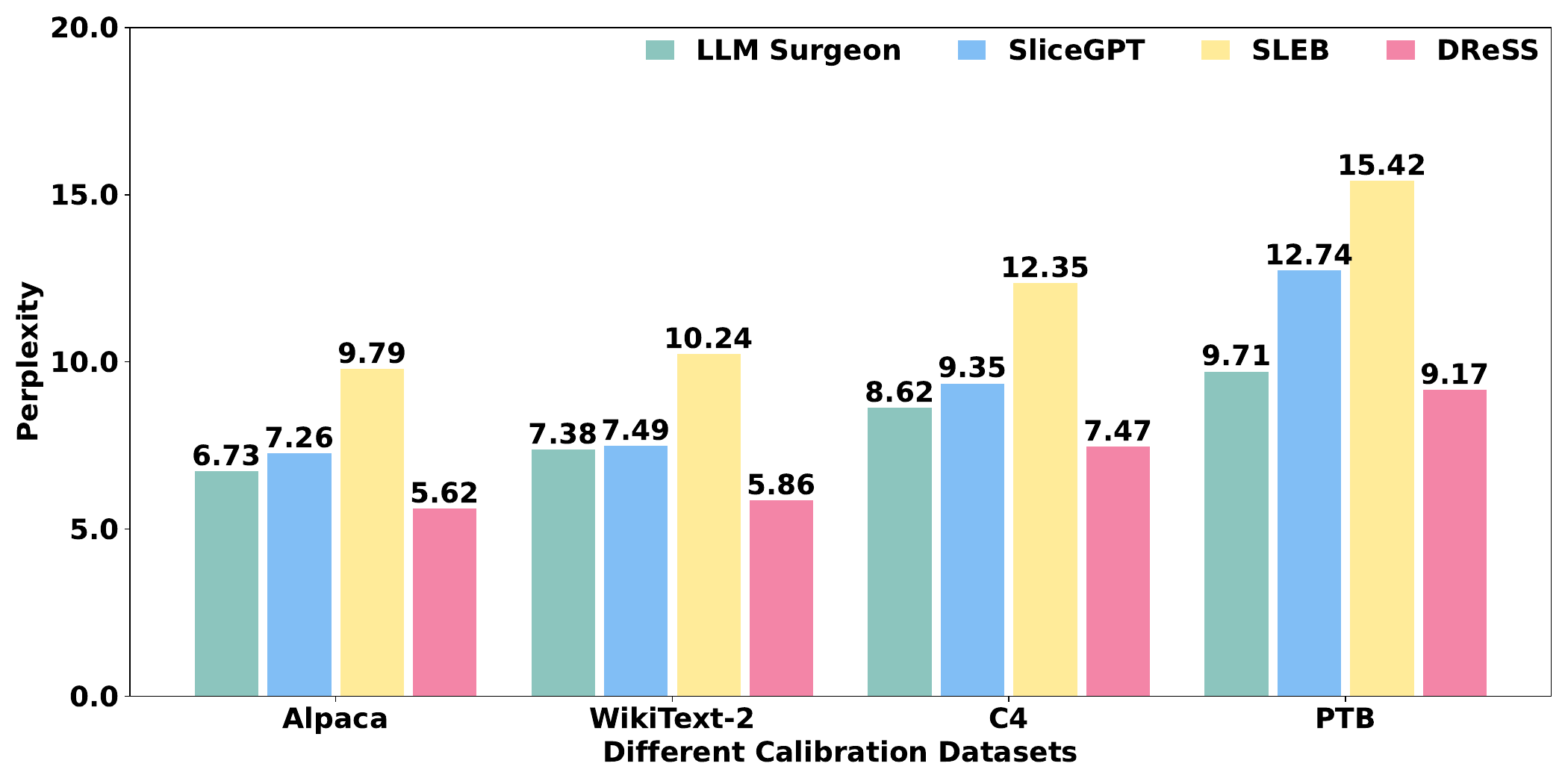}}
\caption{Comparison of perplexity on Wikitext-2 using different calibration datasets at a pruning ratio of 25\% on LLaMA2-7B.}
\label{picture4}
\end{center}
\end{figure}

\begin{figure}[ht]
\begin{center}
\centerline{\includegraphics[width=0.6\textwidth]{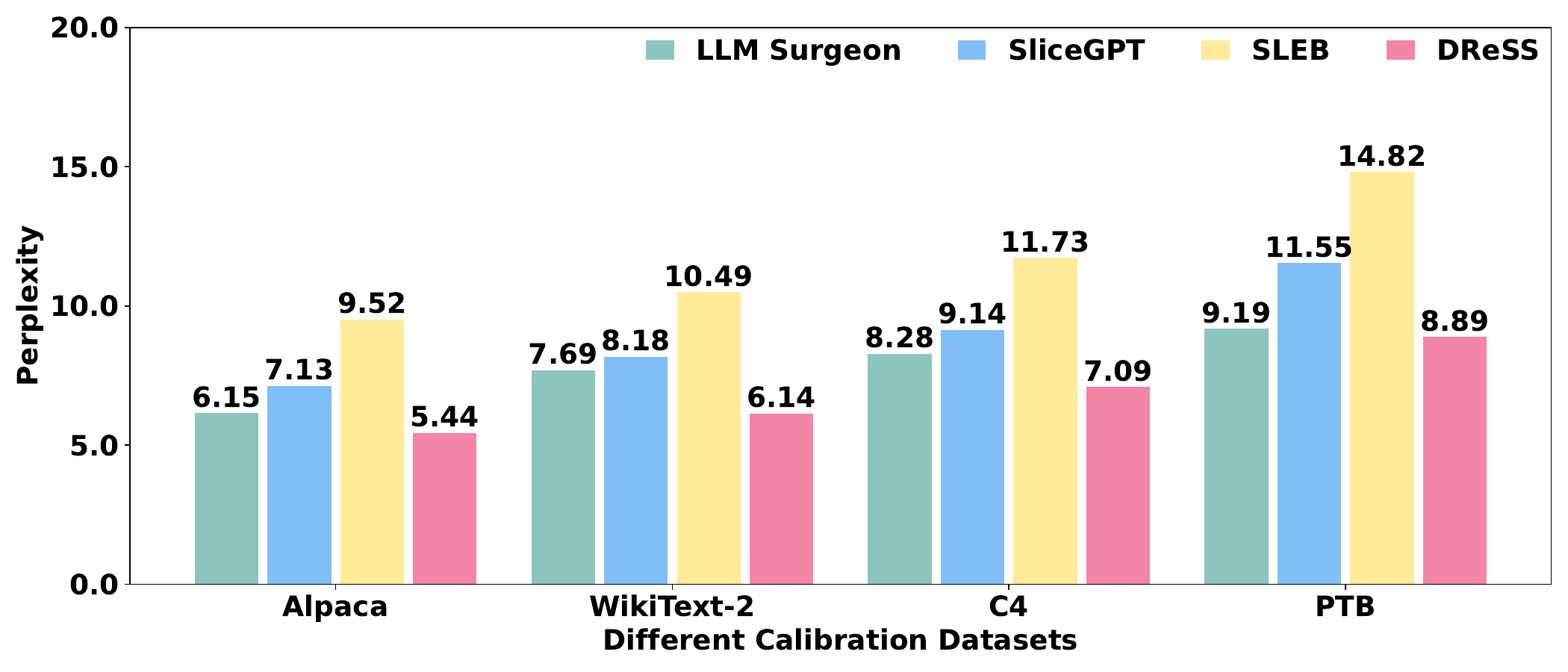}}
\caption{Comparison of perplexity on Alpaca using different calibration datasets at a pruning ratio of 25\% on LLaMA2-7B.}
\label{picture8}
\end{center}
\end{figure}

\begin{figure}[ht]
\begin{center}
\centerline{\includegraphics[width=0.6\textwidth]{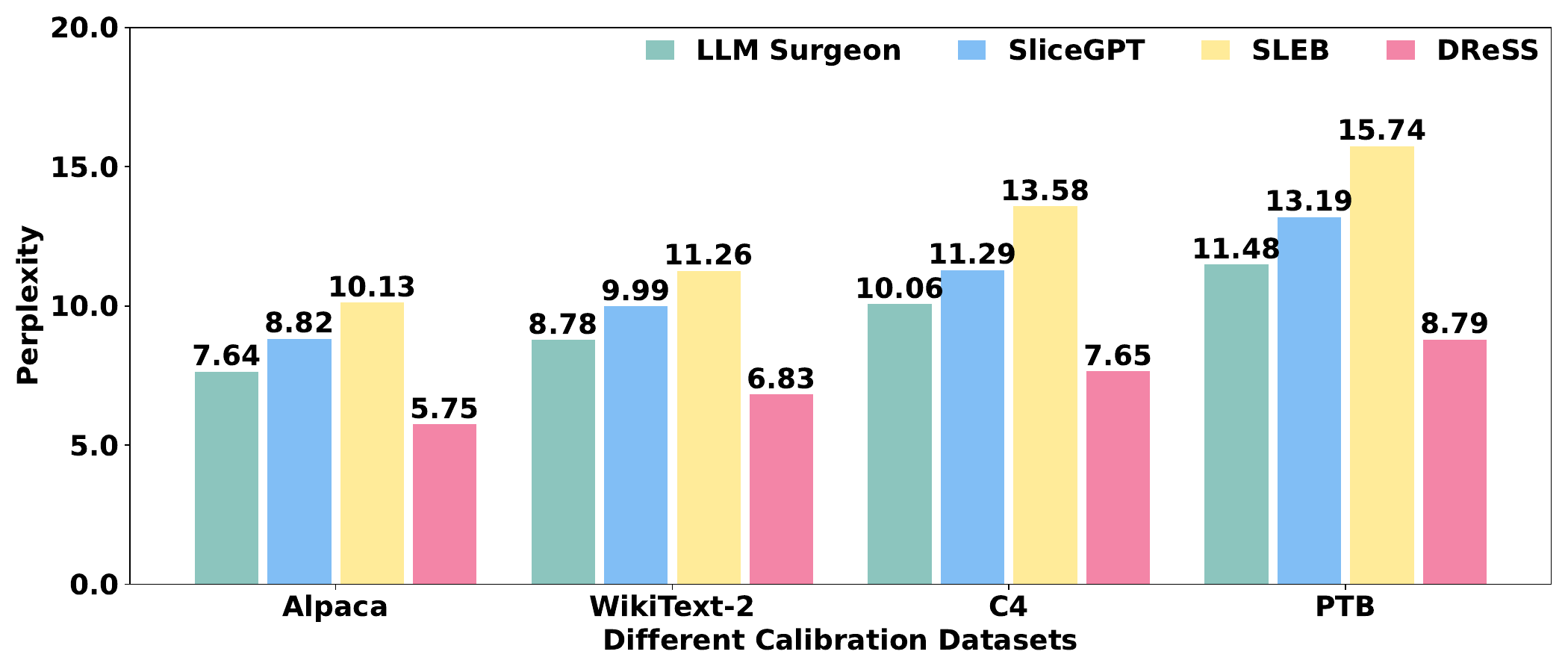}}

\caption{Comparison of perplexity on C4 using different calibration datasets at a pruning ratio of 25\% on LLaMA2-7B.}
\label{picture9}
\end{center}
\end{figure}

\begin{figure}[ht]
\begin{center}
\centerline{\includegraphics[width=0.6\textwidth]{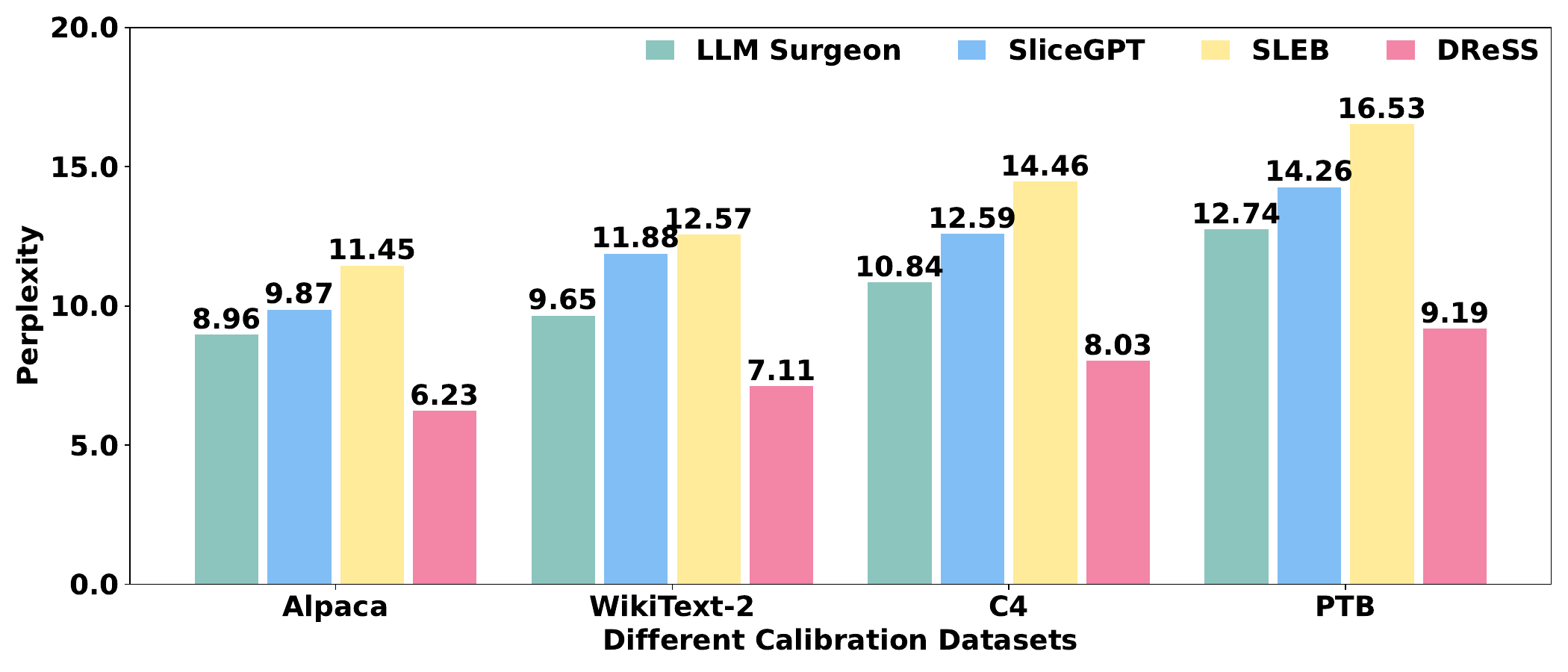}}
\caption{Comparison of perplexity on PTB using different calibration datasets at a pruning ratio of 25\% on LLaMA2-7B.}
\label{picture10}
\end{center}
\end{figure}
\section{Future Work and Limitations}
\label{I}
The ``regularize-then-prune-then-finetune'' paradigm proposed by DReSS is generalizable. In this paper, we only explore pruning at the channel level (i.e., rows or columns of parameter matrices). In future work, we plan to extend the pruning units to entire transformer layers. Using transformer layer as pruning unit is expected to yield more significant acceleration.

\end{document}